# A Dynamic Feedforward Control Strategy for Energy-efficient Building System Operation

Xia Chen [1] Xiaoye Cai [2] Alexander Kümpel [2] Dirk Müller [2] Philipp Geyer [1]

[1] Institute for Sustainable Building Systems, Leibniz University Hannover, Germany
[2] Institute for Energy Efficient Buildings and Indoor Climate, RWTH Aachen University, Germany

*ABSTRACT: The development of current building energy system operation has benefited from: 1. Informational support from the optimal design through simulation or first-principles models; 2. System load and energy prediction through machine learning (ML). Through the literature review, we note that in current control strategies and optimization algorithms, most of them rely on receiving information from real-time feedback or using only predictive signals based on ML data fitting. They do not fully utilize dynamic building information. In other words, embedding dynamic prior knowledge from building system characteristics simultaneously for system control draws less attention. In this context, we propose an engineer-friendly control strategy framework. The framework is integrated with a feedforward loop that embedded a dynamic building environment with leading and lagging system information involved: The simulation combined with system characteristic information is imported to the ML predictive algorithms. ML generates step-ahead information by rolling-window feed-in of simulation output to minimize the errors of its forecasting predecessor in a loop and achieve an overall optimal. We tested it in a case for heating system control with typical control strategies, which shows our framework owns a further energy-saving potential of 15%.*

## 1. INTRODUCTION

The challenge of climate change shapes the shared goal of decarbonatization and sustainability globally. As the building sector consumes up to 30-45% of global energy with the growing trend (Zhong et al., 2021), efforts in building system fine control and management are recognized to contain considerable potential in saving energy (Junker et al., 2018; Mariano-Hernández et al., 2021). In this context, research regarding building control strategy development in Building Energy Management Systems (BEMS) is drawing attention recently. They are classified into two major distinct categories: *Rule-based Controls (RBC)* and *Model Predictive Control (MPC)* (Péan et al., 2019).

RBC with "if-then" trigger rules serves flexibility objectives with substantial performance in load shifting, peak shaving, cost reduction by energy price response, etc. (Bae et al., 2021). The set-point is usually tracked with a traditional PID controller (Madani et al., 2013); however, due to its manual predefined rules/policies, it is less flexible to anticipate optimization with changing external conditions. With the recent boom of artificial intelligence (AI) and the decrease of computation resources cost, MPC surmounts RBC by performance and flexibility for its ability of future behavior estimation, which in the most case benefited by using machine learning (ML) algorithms (Mariano-Hernández et al., 2021; Mendoza-Serrano & Chmielewski, 2014) with its strength of implicit pattern learning ability. In this context, MPC strategies are developed for further improvement via its process-based estimation of defining a specific objective function for minimization (Drgoňa et al., 2020). It is designed to project the future system's behavior and optimize current operation accordingly.

Meanwhile, the flourishing of AI has led to new algorithm adoption of reinforcement learning (RL) based on maximizing the set reward function under environmental constraints (Sutton & Barto, 2018), which has yielded promising results in research (Zhang et al., 2018). RL inherits the strength of ML in well integrating implicit learning factors to deal with external condition changing; however, its modeling process requires reprogramming by different environments, and the model itself is in a black box state, making it less engineer-friendly for deployment and interpretation.

If we carefully inspect the advantages of RL compared to MPC, the former learns implicit relationships of interactive consequences between environment and agent action in a consistent rolling time frame. Whereas MPC strategies based on the ML model merely emphasize the future information input for optimization (Péan et al., 2019). In our opinion, dynamic control is a continuous process requires leading (future) and also lagging (historical) information simultaneously. In this context, prior knowledge embedded with lagging information remains underappreciated. For example, a lagging signal known as autocorrelation in the field of time-domain signal analysis has been shown effective in extracting pattern information (Fulcher, 2018). In this study, we set an objective for research by involving: 1. past behavioral patterns as ML additional inputs via methods, i.e., differencing, and 2. Combined ML with prior knowledge embedded simulation, to improve BEMS performance in the aspect of energy-saving and flexibility.

To address the abovementioned issues, we propose an engineer-friendly, dynamic feedforward control strategy (DFC) framework for BEMS by integrating ML forecasting and physics-based simulation. It carries the advantages of understanding historical/future patterns to ensure performance in system control while operating in advanced energy-efficient behavior.

The remainder of this paper is organized as follows: Section 2 introduces necessary theories and methods for developing forward strategy; Section 3 describes the setup of a case study with benchmark strategies; the result is discussed in Section 4; Section 5 outlines the limitations and future work; Section 6 concludes the paper.

## 2. METHODOLOGY

A general process illustration of the DFC strategy is presented schematically in Figure 1. The core of this strategy is to train an ML model for capturing the dynamic relationship between time-domain features of external weather conditions, indoor temperature changes, and system control states.

**Figure 1:**
*Conceptual process illustration of DFC*

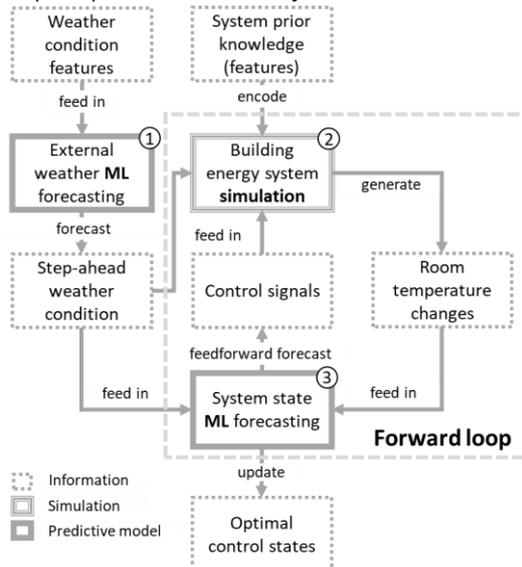

*Note:* DFC strategy consists of two predictive models, including one for external weather prediction (left) and another (bottom) for capturing dynamic behavior between system status and in-/external conditions. It operates in a simulation-involved loop.

### 2.1 Physical-based simulation: Modelica

To complete the desired strategy framework, we apply the building performance simulation (BPS) in Modelica - an open-source modelling language that supports equation-based and object-oriented modelling approaches (Fritzson & Engelson, 1998). The model is based on thermal network models, in which thermal system behaviors are represented with resistances and capacitances of an electrical system based on the analogy of both systems. The German Guideline (VDI 6007-1, 2012) introduces a two-element building model considering internal thermal masses and outer walls, and this model has been evaluated (Lauster et al., 2014). Based on this, (Lauster et al., 2015) introduced the three-elements model by considering one more element for the floor plate. Furthermore, the four-elements model is published in the Modelica library *AixLib* (Müller & et al., 2016): by considering excitations from the exterior wall, interior wall, floor plate, and roof to enable a finer resolution of the dynamic building system behavior. For multiple use cases, this model can be parameterized based on the corresponding databases.

### 2.2 Machine learning model: boosting algorithm

To ensure the ML model recognizes implicit dynamics in time-series data in the state-of-the-art, reviews (Chakraborty & Elzarka, 2019; Chou & Tran, 2018) point out two current prominent advanced algorithm families applied in our domain: neural networks (NNs) and boosting algorithms. Since NNs usually require case-based network structure design, fine-tuning in features of neural nodes, layers, and activation functions to guarantee decent performance, boosting algorithms own the advantage of being "off-the-shelf" without extensive preprocessing or tuning required to perform accurately with generalization flexibility (Tyralis & Papacharalampous, 2021). In this study, we select Light Gradient Boosting Machine (LightGBM) as our ML models. Further insight and an open-source implementation in detail are available in the original paper (Guolin Ke et al., 2017).

### 2.3 Strategy framework

DFC strategy is implemented as step-ahead (stepwise) time-series forecasting, as shown in Figure 2. Dynamics is the keyword of this strategy, which means the objective of this strategy is flexible under different input conditions. For this purpose, a feedforward process is introduced to enable predictive models (ML models) to learn the dynamic relationships between system states and other time-domain features in a range of certain time frames. We designate this system state forecasting model as the target model.

**Figure 2:**
*Semantic illustration of stepwise time-series forecasting with a rolling window*

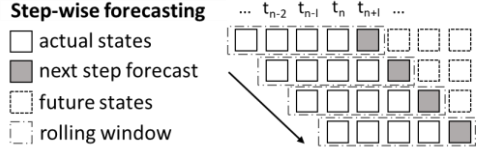

*Note:* $t_n$ stands for the time step. The rolling window means that with the ongoing time, the predive model takes the current state and a certain number of past states as input to forecast the next step. It learns the dynamic relationship in a time-series sequential window.

The core mindset to implementing a feedforward process is to combine the forecasted weather information (Model 1, Figure 1) with the physical-based simulation (Model 2, Figure 1) to obtain the synchronous correspondence between future weather conditions, indoor temperature (setpoint), and system control state (from Model 3, Figure 1). These relationships are fed into the system state forecasting target model (Model 3, Figure 1), thus completing the feedforward loop.

Furthermore, we design the data input structure to ensure that the target model captures the dynamics of the indoor temperature variation in association with time-domain features in an implicit pattern, as presented in Figure 3. The indoor thermodynamic behavior usually has hysteresis effects on external weather conditions and internal load changes, from which the energy-saving potential arises. The target model is designed to access feedforward (leading) and historical (lagging) information within a time-rolling window to make each step of the control strategy optimal with the consideration of consistency.

**Figure 3:**
*Data input/ output structure of the target model*

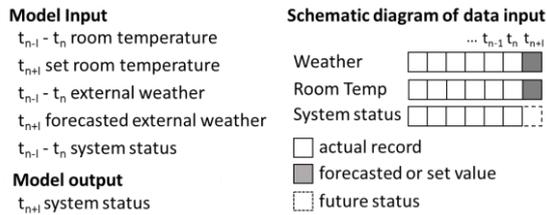

Finally, by combining all components together, the proposed framework pseudo-code is organized as follows:

Input: $\{(x_i, T_i)\}_1^n$, in total $n$ sequential set of building information features $x_i$ (building feature, historical load, weather condition, etc.) with forecasted temperature $T_i$, a differentiable loss function $L(T, F(x))$, number of iterations $M$.
Algorithm:
For $i = 1$ to $n$:
 If the forecasted temperature in the next $M$ steps doesn't fulfill the comfort standard, initialize model with a constant value:

$$F_0(x) = arg\min_\gamma \sum_{i=1}^{M} L(T_i, \gamma) \quad (1)$$

For $m = 1$ to $M$:
a. Feed $F_{m-1}(x)$ into simulation to get the system states output and consequential temperature $(S_m(x_i), F_m(x_i))$.
b. Compute pseudo-residuals: the difference between temperature standard and intermediate predicted temperature:

$$r_m = -\left[\frac{\partial L(T_i, F(x_i))}{\partial F(x_i)}\right]_{F(x)=F_{m-1}(x)} \quad (2)$$

c. Fit a regression tree to the $r_m$ values and create terminal regions $R_{j,m}$ for $j = 1, \dots, J_m$
d. Compute multiplier $\gamma_{j,m}$ by:
$$\gamma_{j,m} = arg\min_\gamma \sum_{x_i \in R_{j,m}} L(T_i, F_{m-1}(x_i) + \gamma) \quad (3)$$
e. Update the model with forecasted temperature:
$$F_m(x) = F_{m-1}(x) + \nu \sum_{j=1}^{J_m} \gamma_{j,m} \quad (4)$$

Output: control state sequential $\{(S_M(x_i))\}_{i-1}^n$

In summary, the novelty, and the performance improvement of the DFC strategy come from the embedding of physical information dynamically into the objective function in the data-driven process, which means that the DFC enable:
- Recognition of encoded building physical information from simulation output data by the ML model.
- Combining historical indoor temperature changes and states, training the relationship between future weather and system states through weather leading information combined with simulation.
- Dynamic pattern learning from lagging input of external weather conditions, indoor temperature, and system states information, thus optimizing system control states.

## 3. CASE DESCRIPTION
In the case study, we simulate a typical office building in Modelica with the standard of the passive house (Tian et al., 2018). The model handles solar radiation and internal gains, which contain the emitted heat from occupants, equipment, and lighting. The database parameterizing the office building was validated and has been used in several works (Mork et al., 2022; Niederau et al., 2021). The office area is 1675 m² (parameters in Table 1), equipped with space heaters and a Viessman heat pump with nominal power of 18,5 kW (scheme in Figure 4). Models and data are accessible in *AixLib*.

**Figure 4:**
*The system scheme of the simulation model*

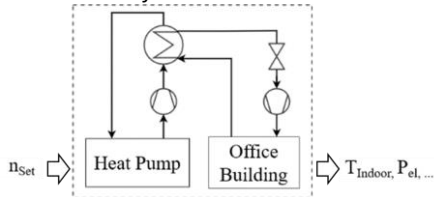

Note: $n_{set}$: Relative rotational speed of compressor in the heat pump; $T_{Indoor}$: indoor temperature; $P_{el}$: electrical power.

**Table 1:**
*Parameters of the office building simulation with passive house standard*

| Parameter | Value | Unit |
|---|---|---|
| Resistances of exterior walls | 1.41·10⁻⁴ | K/W |
| Heat capacities of exterior walls | 4.93·10⁸ | J/K |
| Resistances of floor plate | 1·10⁻³ | K/W |
| Resistances of roof | 1·10⁻³ | K/W |
| Resistances of interior wall | 1.3·10⁻⁴ | K/W |

In this simulation, we define two rules for the comfort objective in temperature: *IF day (7 AM – 6 PM), THEN the temperature setpoint is 21°C; IF night, THEN the temperature setpoint is 19°C.* The simulation is set under the weather condition of Aachen, Germany. The simulated behavior of indoor temperature is presented in Figure 5.

**Figure 5:**
*Indoor temperature behavior without control*

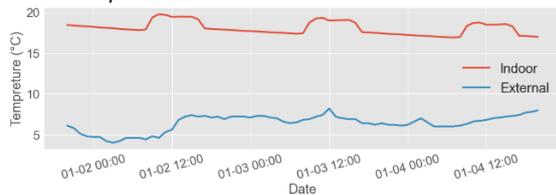

Based on the set scenario, we implement the desired feedforward strategy (DFC) mentioned in section 2 for fulfilling the comfort objective. To evaluate DFC, we compare it with two Reference Control (RC1 and RC2): RC1 is based on a proportional-integral-derivative controller (PID). A PID employs a responsive correction to a control function in real-time and hence the PID-based control aims at reducing the difference between the set indoor temperature and the measured value. RC2 features predictive control, which uses leading signals by simulation: When the indoor temperature is below the set temperature, by cumulative increasing *nSet* in 0.05 step length until the room temperature fulfills the comfort objective. All predictive strategies are applied in 10-minute time-step granularity.

RC2 and DFC strategies require the feed-in of future signals for process estimation. We used historical Aachen weather with features presented in Table 2 with lagging signals of features, and trained the ML model for one-step-ahead (as in Figure 2) external weather condition prediction with a rolling window. The result (plotted in Figure 6) shows high accuracy (0.9896 coefficient of determination, or R squared). The one step ahead output of air temperature is then merged into MPC strategy and DFC strategy model training and control (as in Figure 3).

**Table 2:**
*Weather features*

| Input feature | Description |
|---|---|
| *temp* | Air temperature (°C) |
| *dew* | Dew point (°C) |
| *hum* | Relative humidity (%) |
| *pres* | Atmospheric pressure (hPa) |
| *winds* | Wind speed (m/s) |

**Figure 6:**
*Weather forecasting result*

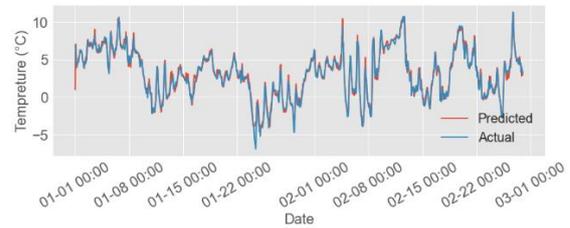

## 4. RESULTS

Table 3 shows a quantitative comparison of energy consumption within three strategies, while Figure 7 plots them in *nSet, indoor temperature,* and *heat pump coefficient of performance (COP)*, respectively.

**Figure 7:**
*Comparison of performance results in different strategies.*

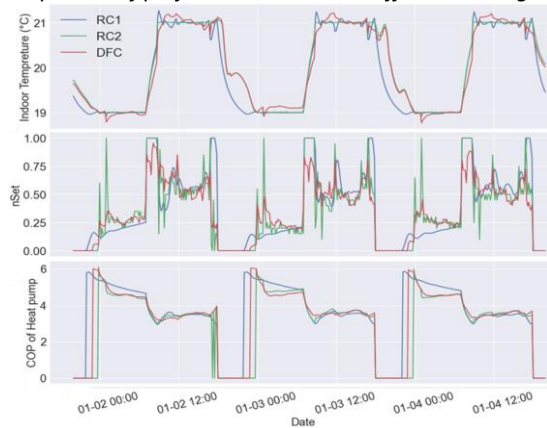

**Table 3:**
*Comparison of electrical energy consumption in different strategies*

| Strategy | Energy consumption |
|---|---|
| RC1 (PID-based) | 0.305 kWh/day per m$^2$ |
| RC2 (predictive) | 0.273 kWh/day per m$^2$ |
| **DFC** | **0.251** kWh/day per m$^2$ |

*Note:* Calculated based on the daily average from Figure 7.

The quantitative result shows that compared to RC2 and DFC, RC1 performs the least efficient control. We conclude the reason from the subfigure of *nSet* in Figure 7: With no future signal feed-in for process optimization, the PID-based RC1 causes oscillation at the temperature set point against temperature changes. Furthermore, because of its real-time response to temperature changes, a stable temperature curve is observed during daylight. The cost of this fine regulation results in a large amount of energy consumption by keeping the compressor running with oscillation to maintain the temperature above the minimum setting.

Compared to the RC1 control, RC2 performs in a more promising behavior with 10.5% lower energy consumption. The difference is that RC2 utilizes predictive leading signals of indoor temperature to regulate the energy system in less oscillation behavior and total energy consumption. However, RC2 has more spikes in regulation *(nSet)*.

The DFC strategy fulfills the comfort objective with a further advantage in energy-efficient control: 17.7% reduced energy consumption. Due to combined information from lagging and leading signals, DFC comprehends dynamic time-domain behaviors well and produces smoother control states with less overshooting. In return, the temperature curve is not strictly aligned with the standard with ±0.5°C fluctuation. The reason for this phenomenon is two-fold: the unpunished temperature overshooting behavior in the algorithm, and the accumulated inaccuracy of signals because of two ML models involved DFC strategy, including one for external weather prediction (Model 1 in Figure 1), and another for capturing dynamic behavior between weather, indoor temperature, and energy system states (Model 3 in Figure 1).

The highlight of the DFC is a new control pattern observed in Figure 7: Before the set temperature changes in the morning, we observed the DFC start the regulation in the time between RC1 and RC2. It means that the DFC strategy learns dynamic relationships among time-domain features in-depth by rolling stepwise training from physics-informed simulation. Furthermore, compared to RC1 or RC2, the DFC strategy tends to decompose a substantial change (upward or downward spike) into two to multiple small spikes. By taking both lagging and leading information into account, the strategy raises small spikes to preheat the room with continuous control sequence to ensure indoor temperature changes within the set frame.

## 5. DISCUSSION

This section discusses the strategy adaptability, limitation, and future development prospect.

DFC strategy is designed to fill the gap in the current MPC-based strategy by combining the physical-based simulation and data-driven models. The general framework design aims to fully utilize the domain knowledge (building characteristics) encoded in the simulation model and the implicit pattern (time-domain dynamic relationships) recognition ability of ML. *More specifically, it makes the prior-knowledge validation accessible in the loop of the data-driven optimizing process.* This mindset is applicable in general engineering domains. Within the scope of BEMS control optimization, we've proved its effectiveness with good interpretability. In this aspect, we encourage investigations in different scenarios to validate its performance robustness.

In this study, the setup scenario with external weather conditions, indoor temperature, and system status in the framework is relatively simple. DFC has the same potential as MPC to be extended in complex scenarios to achieve control strategies under different objectives, such as minimization of cost, $CO_2$ intensity, non-renewable primary energy, etc. by customizing the objective function.

Although the strategy shows advantages from an energy-saving perspective, the process of model training requires efforts in case-based simulation modeling. In this context, the trade-off investigation between modeling detail and performance is necessary. Future works in constructing a general simulation framework for a certain type of buildings to fit a broader range of application scenarios are meaningful.

## 6. CONCLUSION

In this study, based on current limitations in BEMS control strategies, we proposed a dynamic feedforward strategy to achieve stepwise rolling predictive system control, which contains two novelties: *integration of physical-based simulation with engineer-friendly ML models*, and *utilization of leading and lagging indicators with physics-informed information*, to conduct consistent and energy-efficient control signals. The designed case shows that the proposed strategy outperforms PID-controller and MPC-based reference strategy in terms of BEMS control and optimization. Under the general trend of digitalization and data-driven model application, the domain knowledge integration from the perspective of informed ML methods should raise further attention within our community to contribute to the objective of sustainable development.

**ACKNOWLEDGEMENTS**
We thank Yusheng, Chen for useful discussions. We acknowledge the German Research Foundation (DFG) support for funding the project under grant GE 1652/3-2 in the Researcher Unit FOR 2363.